\documentclass[conference]{IEEEtran}
\IEEEoverridecommandlockouts
\usepackage{cite}
\usepackage{amsmath,amssymb,amsfonts}
\usepackage{algorithmic}
\usepackage{graphicx}
\usepackage{textcomp}
\usepackage[hyphens]{url}
\usepackage[table,xcdraw]{xcolor}
\usepackage{multirow}
\usepackage{subfigure}
\usepackage{threeparttable}
\def\BibTeX{{\rm B\kern-.05em{\sc i\kern-.025em b}\kern-.08em
    T\kern-.1667em\lower.7ex\hbox{E}\kern-.125emX}}
\begin{document}

\title{Diagonalwise Refactorization: An Efficient\\ Training Method for Depthwise Convolutions
}

\def\todo#1{\textcolor{red}{#1}}

\author{
\IEEEauthorblockN{Zheng Qin, Zhaoning Zhang$^*$\thanks{$^*$Corresponding author.}, Dongsheng Li, Yiming Zhang, Yuxing Peng}
\IEEEauthorblockA{\textit{Science and Technology on Parallel and Distributed Laboratory} \\
\textit{National University of Defense Technology}\\
Changsha, China\\
qinzheng12@nudt.edu.cn; zzningxp@gmail.com; lds1201@163.com; sdiris@gmail.com; pengyuxing@nudt.edu.cn}
}

\hyphenation{networks methods para-meters appears op-tical net-works semi-conduc-tor store stor-age snap-shot space reads among Ursa exits scal-able uses sys-tems IOPS dif-ferent GFS HDFS FS gen-erally SCSI NFS HBase SSD slightly stored two anony-mized speak-ing lineariza-bility server elim-inate states pro-cesses signifi-cantly sin-gle OCFS blocks par-allelism parallel-ism returns pre-senting present-ing stand-ard stores IOPS servers appends scala-bility pri-mary mode respec-tively re-spectively sat-isfy UPSs CPUs avail-ability availa-bility availabil-ity clouds usu-ally against re-quests inde-pendently namely storing Meituan evolv-ing since ob-ject tests BS mounted under-loaded consis-tency status jour-nals place-ment statis-tics exclu-sively meta-data mainly collabora-tively collabo-ratively FDS focus nodes remote conven-tional strata single enough daily require small sequen-tial Theore-tically coroutine acce-leration accele-ration effi-cient matrix vectors imple-mentations implemen-tations implementa-tions larger Mobile-Nets GPUs com-puted cuDNN hyper-parameters hyperpara-meters Con-sequently Conse-quently app-lications appli-cations applica-tions adopted GPU accele-rating acce-lerating Tensor-Flow results frame-works convo-lutions con-volutions convolu-tions convo-lution con-volution convolu-tion threads sig-nificant signi-ficant signifi-cant ii iii GEMM aca-demic acade-mic its weights algo-rithms uti-lize also rela-tively redun-dancy imple-mentation implemen-tation implementa-tion Mobile-Net multi-plier fewer res-pectively respec-tively kernel table uti-lizes every in-creases diag-onalwise Multiplications}

\hyphenpenalty=1000


\maketitle

\begin{abstract}
Depthwise convolutions
provide significant performance benefits
owing to the reduction in both parameters and mult-adds.
However, training depthwise convolution layers with GPUs is slow in current deep learning frameworks
because their implementations cannot fully utilize the GPU capacity.
To address this problem,
in this paper we present an efficient method (called \textit{diagonalwise refactorization})
for accelerating the training of depthwise convolution layers.
Our key idea is to
rearrange the weight vectors of a depthwise convolution into a large diagonal weight matrix
so as to convert the depthwise convolution into one single standard convolution,
which is well supported by the cuDNN library that is highly-optimized for GPU computations.
We have implemented our training method in five popular deep learning frameworks.
Evaluation results show that
our proposed method gains $15.4\times$ training speedup on Darknet, $8.4\times$ on Caffe, $5.4\times$ on PyTorch, $3.5\times$ on MXNet, and $1.4\times$ on TensorFlow,
compared to their original implementations of depthwise convolutions.
\end{abstract}


\section{Introduction}

Deep convolutional neural networks (ConvNets) \cite{krizhevsky2012imagenet,simonyan2014very,szegedy2015going,he2016deep,ren2015faster,badrinarayanan2017segnet} have recently become increasingly important for computer vision applications.
However,
standard deep ConvNets suffer from high computational cost
due to their high numbers of parameters and mult-add operations.

MobileNets \cite{howard2017mobilenets} uses \textit{depthwise separable convolutions},
which factorize a standard convolution into a depthwise convolution and a pointwise ($1 \times 1$) convolution,
to effectively reduce the numbers of both parameters and mult-add operations.
Xception \cite{chollet2016xception} leverages depthwise separable convolutions
to improve its classification performance.
However,
as reported in \cite{zhang2017shufflenet} and \cite{wu2017shift},
depthwise convolutions have a low computation \textit{vs.} memory access rate,
which means memory access takes more execution time than computation and it is more difficult to implement depthwise convolutions as efficient as computation-intensive layers like standard convolutions.
This makes training depthwise convolution layers with GPUs very slow in current deep learning frameworks such as Caffe \cite{jia2014caffe}, PyTorch \cite{collobert2011torch7}, MXNet \cite{chen2015mxnet} and TensorFlow \cite{abadi2016tensorflow},
mainly because their implementations of depthwise convolutions cannot fully utilize the GPU capacity.

Caffe, PyTorch and MXNet
implement depthwise convolutions
by performing the standard convolution \textit{channel-by-channel}.
This method simply launches a CUDA kernel or cuDNN function for each of the input channels,
and applies no inter-channel optimizations such as filter combination.
Consequently,
the number of threads launched for each standard convolution is small
and the utilization of GPU cores is very low.
For example,
although depthwise convolutions have only about 3\% of the mult-adds and 1\% of the parameters
when training a MobileNet \cite{howard2017mobilenets},
they spend over 82\% of the overall training time  on Caffe which is much higher than any other layer types,
as shown in our evaluation (Table \ref{table:layer-type-1}).

Different from the channel-by-channel method,
TensorFlow
adopts the \textit{specialized kernel} method
which
implements depthwise convolutions
by designing a specialized CUDA kernel
and computing all the input channels in the single kernel.
This method is more efficient in training depthwise convolution layers
because it exploits the inter-channel parallelism.
However,
the specialized kernel method prevents TensorFlow from leveraging the cuDNN library \cite{chetlur2014cudnn}
with the algorithm-level and microarchitecture-level optimizations,
which are vital for high-performance GPU computations.

\begin{table}[t]
\centering
\begin{threeparttable}[t]
\caption{Ratios of mult-adds, parameters, and training time of different layer types for MobileNets on Caffe.}
\label{table:layer-type-1}
\begin{tabular}{|c|c|c|c|}
\hline
\textbf{Type} & \textbf{Mult-Adds} & \textbf{Parameters} & \textbf{Training Time} \\ \hline
Conv $1 \times 1$ & 94.86\% & 74.59\% & 16.39\% \\ \hline
Conv DW $3 \times 3$ & 3.06\% & 1.06\% & 82.86\% \\ \hline
Conv $3 \times 3$ & 1.19\% & 0.02\% & 0.72\% \\ \hline
Fully Connected & 0.18\% & 24.33\% & 0.03\% \\ \hline
\end{tabular}
Conv DW: depthwise convolution layer.
\end{threeparttable}
\end{table}

This paper presents \textit{diagonalwise refactorization},
an efficient method for accelerating the training of depthwise convolution layers.
First,
the weight vectors (filters) of the input channels are rearranged into a diagonal matrix
to construct one single large filter.
Then,
the depthwise convolution is computed as a standard convolution with the large filter,
which supports to leverage the cuDNN library to accelerate the computation.
We further adopt a grouping mechanism for convolutions with large numbers of input channels,
where
the channels are divided into several groups
and the diagonalwise refactorization is performed for each group.
By combining all filters into a large one,
our method could exploit the inter-channel parallelism
to utilize the GPU capability more efficiently.
By supporting the cuDNN library,
our method could directly enjoy its algorithm-level and microarchitecture-level optimizations.

We have implemented our method
on five popular frameworks
including Darknet \cite{darknet13}, Caffe, PyTorch, MXNet, and TensorFlow.
Evaluation results show that
our method gains $15.4\times$ speedup on Darknet, $8.4\times$ on Caffe, $5.4\times$ on PyTorch, $3.5\times$ on MXNet, and $1.4\times$ on TensorFlow, when training a standard MobileNet,
compared to their original implementations of depthwise convolutions.
We conduct extensive experiments on different MobileNet hyper-parameters
including shallow models, width multiplier and resolution multiplier,
and perform detailed analysis on the layer-by-layer training time.
Code has been made publicly available at \url{https://github.com/clavichord93}\footnote{Code in Caffe, PyTorch and TensorFlow has been made publicly available at \url{https://github.com/clavichord93/diagonalwise-refactorization-caffe}, \url{https://github.com/clavichord93/diagonalwise-refactorization-pytorch} and \url{https://github.com/clavichord93/diagonalwise-refactorization-tensorflow}.}.

The contribution of this paper is summarized as follows.
\begin{enumerate}
  \item We propose a novel method (diagonalwise refactorization) that effectively accelerates the training of depthwise convolutions.
  \item We implement our method on five popular frameworks
  and provide detailed performance comparison and \mbox{analysis}.
  \item We discuss the extensibility of our method
  and show that it could be adopted in the training of many acceleration techniques
  such as pruning and group convolutions.
\end{enumerate}

\section{Related Work}

Many techniques have been proposed to compress existing ConvNets.
Network pruning \cite{han2015learning,han2015deep,he2017channel,liu2017learning} accelerates the inference of networks by reducing spatial, connection and channel redundancy.
Parameter quantization \cite{courbariaux2015binaryconnect,hubara2016binarized,rastegari2016xnor,zhou2016dorefa,zhou2017incremental,chen2017fxpnet} trains deep ConvNets directly with binary weights and gains significant acceleration.
Tensor decomposition \cite{denton2014exploiting,lebedev2014speeding,zhang2016accelerating} adopts a low-rank approximation to original convolution filters to reduce parameters.

Layer factorization
has been proposed to build lightweight networks.
SqueezeNet \cite{iandola2016squeezenet} proposes fire modules which mix $1 \times 1$ and $3 \times 3$ convolutions and achieves AlexNet-level accuracy with $50 \times$ fewer parameters.
MobileNet \cite{howard2017mobilenets} replaces standard convolutions with depthwise separable convolutions \cite{chollet2016xception} and provides competitive accuracy with state-of-the-art networks.
ShuffleNet \cite{zhang2017shufflenet} further applies group convolutions in depthwise covolutions with channel shuffle operations.

\begin{figure}[t]
\centering
\includegraphics[width=0.45\textwidth]{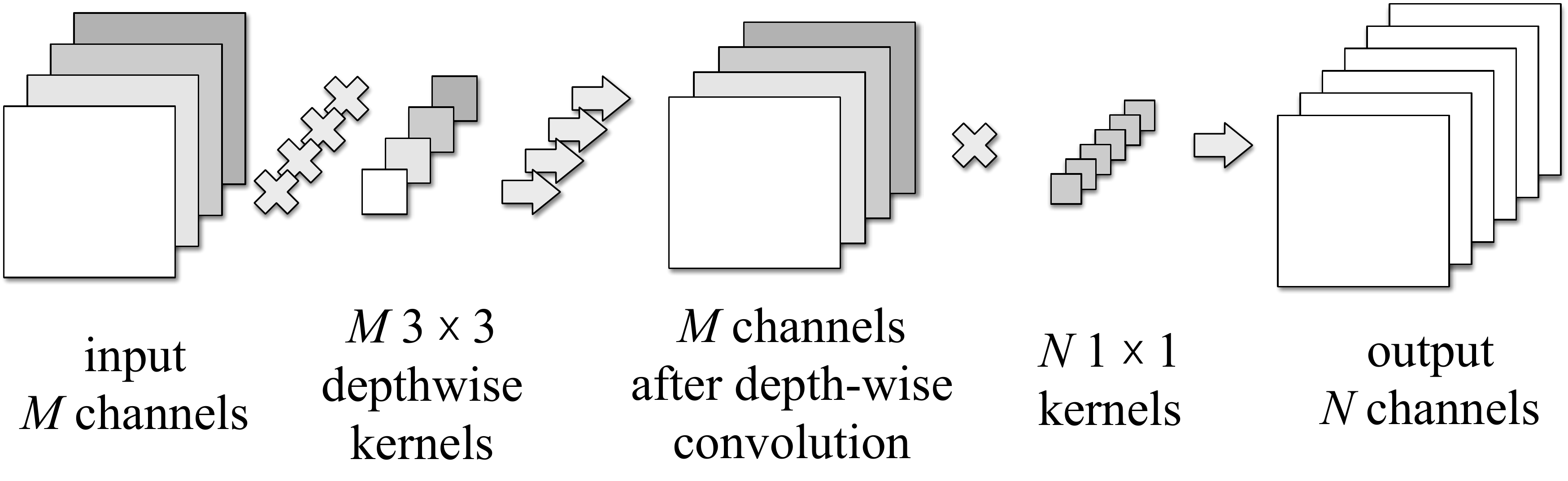}
\caption{A depthwise separable convolution is composed of a depthwise convolution and a pointwise convolution.}
\label{figure:depthwise-separable-convolution}
\end{figure}

A depthwise separable convolution is a combination of a depthwise convolution and a pointwise convolution.
As shown in Figure \ref{figure:depthwise-separable-convolution},
a depthwise convolution filter (kernel) is applied to one input channel with its own set of weights.
For an $M$-channel input feature map, a depthwise convolution creates an $M$-channel output feature map.
Depthwise separable convolutions achieve a significant reduction in parameters and mult-add operations,
and thus effectively accelerate the computation of ConvNets.
For example,
MobileNets perform very fast inference on mobile devices \cite{howard2017mobilenets,huang2016speed,hollemans2017mobilenets}.
We also achieve real-time single-category detection with a 0.375-MobileNet-416 model on i7 CPU and a 0.375-MobileNet-128 model\footnote{Only the first 11 layers of the standard MobileNet is used. 0.375 indicates the width multiplier while 416 and 128 indicate the resolution multiplier.} on iMx6 ARM with NEON acceleration, in the YOLOv2 \cite{redmon2016yolo9000} detection framework.

\begin{figure}[t]
\centering
\includegraphics[width=0.47\textwidth]{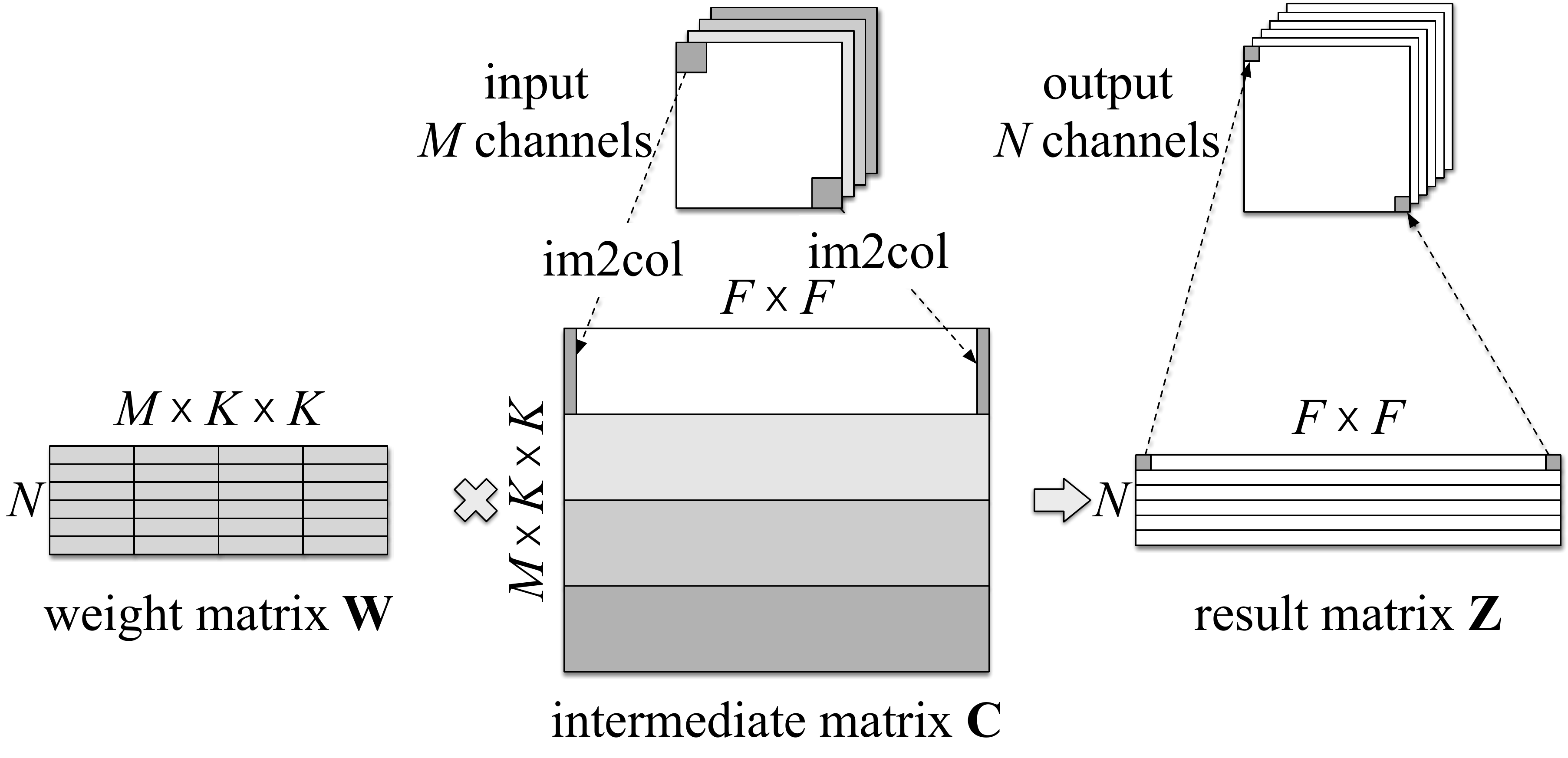}
\caption{Matrix manipulation in standard convolutions. The intermediate matrix $\mathbf{C}$ is first obtained by an im2col operation. Then a multiplication between the weight matrix $\mathbf{W}$ and $\mathbf{C}$ produces the result matrix $\mathbf{Z}.$}
\label{figure:matrix-standard}
\end{figure}

Currently there are two methods (\textit{channel-by-channel} and \textit{specialized kernel})
for implementing depthwise convolutions,
both of which are based on the \textit{standard convolutions}.

\noindent
\textbf{Standard convolutions}.
Assume that the numbers of input and output channels are respectively $M$ and $N$,
the size of the feature map is $F \times F$, and the size of the convolutional kernels is $K \times K$.
A standard convolution is implemented in two steps,
as shown in Figure \ref{figure:matrix-standard}.
The first step is an im2col operation,
where the input feature map $\mathbf{X}_{M \times F \times F}$ is rearranged
into an intermediate matrix $\mathbf{C}_{(M \cdot K \cdot K) \times (F \cdot F)}$,
and every block to be convolved in $\mathbf{X}$ is rearranged into a column of $\mathbf{C}$.
The second step is a matrix multiplication between the convolution weight matrix $\mathbf{W}_{N \times (M \cdot K \cdot K)}$ and the intermediate matrix $\mathbf{C}$.
The result is a matrix $\mathbf{Z}_{N \times (F \cdot F)}$,
each row of which represents a single channel of the output feature map.

\begin{figure}[t]
\centering
\includegraphics[width=0.47\textwidth]{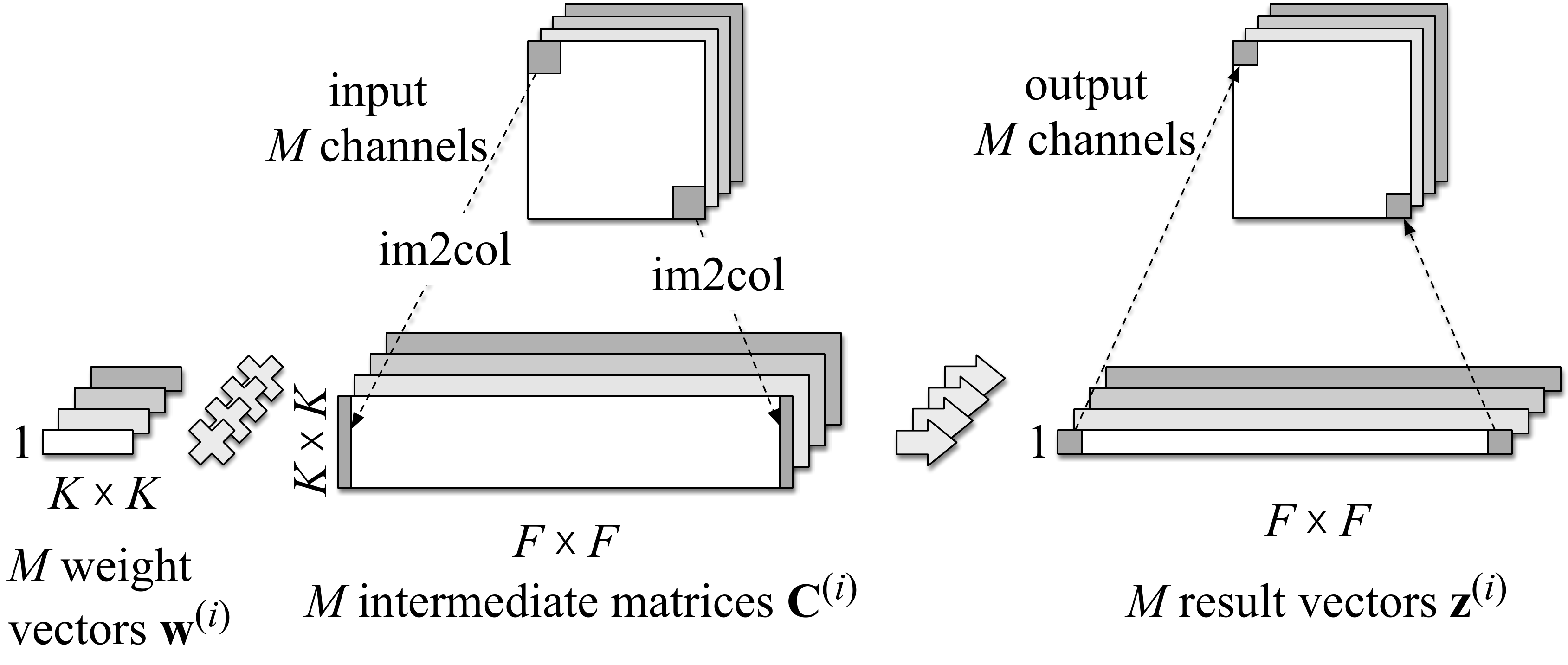}
\caption{Matrix manipulation in the channel-by-channel method. Multiplications between weight vectors $\mathbf{w}^{(i)}$ and intermediate matrices $\mathbf{C}^{(i)}$ produce the result vectors $\mathbf{z}^{(i)}$.
}
\label{figure:matrix-channel-wise}
\end{figure}

\noindent
\textbf{Channel-by-channel method}.
As shown in Figure \ref{figure:matrix-channel-wise},
this method simply performs the standard convolution for each of the input channels.
$M$ intermediate matrices $\mathbf{C}^{(i)}_{(K \cdot K) \times (F \cdot F)}$ are first generated using an im2col operation.
Then $M$ multiplications between the weight vectors $\mathbf{w}^{(i)}_{K \times K}$ and the intermediate matrices $\mathbf{C}^{(i)}$ are performed.
Finally, tiling the $M$ result vectors $\mathbf{z}^{(i)}_{F \times F}$ generates the result matrix $\mathbf{Z}$.
The channel-by-channel method is adopted in Caffe, PyTorch and MXNet.
Caffe only supports to leverage the general matrix multiply (GEMM) operations,
but not cuDNN,
for channel-by-channel convolutions,
while PyTorch and MXNet support to use cuDNN
which offers a remarkable performance improvement compared to Caffe (as shown in Table \ref{table:overall-results}).

\noindent
\textbf{Specialized kernel method}.
This method does not explicitly generate the intermediate matrix $\mathbf{C}$
and designs its own specialized CUDA kernel for implementing depthwise convolutions
instead of performing standard convolutions channel-by-channel with GEMM or cuDNN.
In a CUDA thread,
a $K \times K$ block from the input feature map $\mathbf{X}$ is convolved with the weights $\mathbf{w}^{(i)}$ of the same channel to compute one pixel in the output feature map $\mathbf{Z}$.
The specialized kernel method is adopted in TensorFlow,
which performs its own hand optimizations for GPU's shared memory and caches.

\section{Design}

In this section,
we first introduce the diagonalwise refactorization method,
and then describe the grouping mechanism for large numbers of input channels.
At last we analyze the advantage of our method over previous proposals.

\subsection{Diagonalwise Refactorization}

\begin{figure}[t]
\centering
\includegraphics[width=0.47\textwidth]{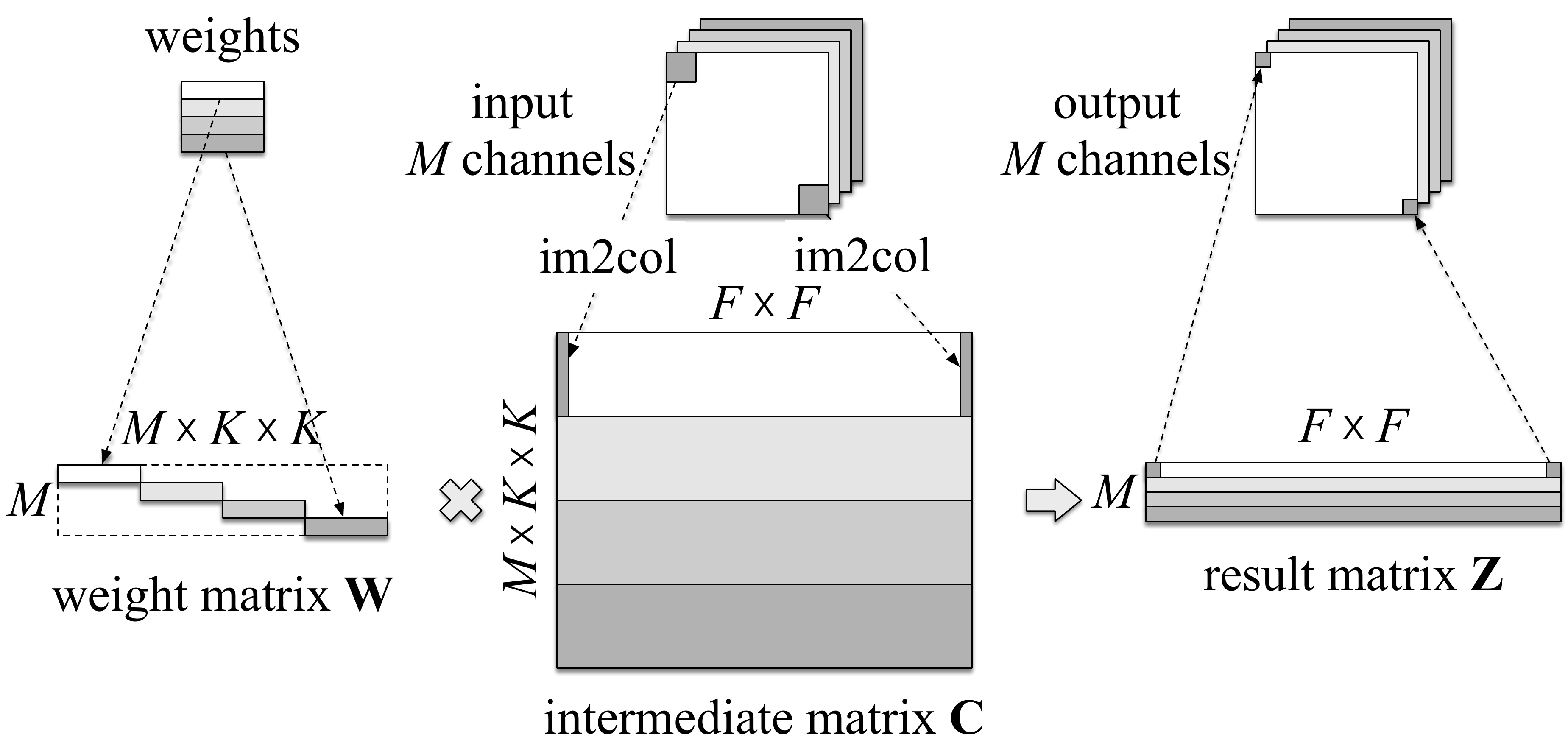}
\caption{Matrix manipulation in a 4-channel diagonalwise group. The weight vectors $\mathbf{w}^{(i)}$ are first rearranged into a large weight matrix $\mathbf{W}$. Then the multiplication of $\mathbf{W}$ and the intermediate matrix $\mathbf{C}$ produces the result matrix $\mathbf{Z}$.}
\label{figure:matrix-diagonalwise}
\end{figure}

Consider a depthwise convolution with $M$ input channels.
In normal depthwise convolutions, a single filter is a vector $\mathbf{w}^{(i)}$ of length $K \times K$.
The convolution operation is the multiplication of a weight vector $\mathbf{w}^{(i)}$ and the intermediate matrix $\mathbf{C}^{(i)}$ of the same input channel.
The depthwise convolution is composed of $M$ vector-matrix multiplications.
In diagonalwise refactorization, we convert a depthwise convolution into a standard convolution.
The $M$ weight vectors $\mathbf{w}^{(i)}$ are rearranged into a large weight matrix $\mathbf{W}_{M \times (M \cdot K \cdot K)}$ on the diagonal positions and all other positions are set to 0.
The im2col matrices $\mathbf{C}^{(i)}$ are tiled from top to bottom and form a large intermediate matrix $\mathbf{C}_{(M \cdot K \cdot K) \times (F \cdot F)}$, which is the same as in a standard convolution.
Figure \ref{figure:matrix-diagonalwise} shows how we rearrange the depthwise weight vectors into a large weight matrix.
During backward propagation,
the gradients from the top layer are passed only to the diagonal weights and all other positions remain 0.

This conversion can be expressed as an $M$-channel standard convolution preceded by a multiplication of the weight matrix $\mathbf{W}$ and a constant mask matrix $\mathbf{A}_{M \times (M \cdot K \cdot K)}$ where

\begin{equation}
  \mathbf{W}=\left[
  \begin{array}{cccc}
    \mathbf{w}^{(1)} & & & \\
    & \mathbf{w}^{(2)} & & \\
    & & \ddots & \\
    & & & \mathbf{w}^{(M)} \\
  \end{array}\right],
\end{equation}

\begin{equation}
  \mathbf{A}=\left[
  \begin{array}{cccc}
    \mathbf{1}_{1 \times (K \cdot K)} & & & \\
    & \mathbf{1}_{1 \times (K \cdot K)} & & \\
    & & \ddots & \\
    & & & \mathbf{1}_{1 \times (K \cdot K)} \\
  \end{array}\right].
\end{equation}

\noindent
The depthwise convolution could be written as
\begin{equation}
\begin{split}
  \hat{\mathbf{W}} & = \mathbf{W} \odot \mathbf{A} \\
  \mathbf{Z} & = \hat{\mathbf{W}} \otimes \mathbf{X}
\end{split}
\end{equation}

\noindent
where $\mathbf{X}$ is the input feature map, $\mathbf{Z}$ is the output feature map, $\odot$ represents elementwise multiplication and $\otimes$ represents convolution.
With the mask matrix $\mathbf{A}$,
redundant weights are filtered out and only depthwise weights are used for convolution.
During backward propagation, the gradients of the weight matrix are
\begin{equation}
\frac{\partial \mathbf{Z}}{\partial \mathbf{W}} = \frac{\partial \mathbf{Z}}{\partial \hat{\mathbf{W}}} \cdot \frac{\partial \hat{\mathbf{W}}}{\partial \mathbf{W}} = \frac{\partial \mathbf{Z}}{\partial \hat{\mathbf{W}}} \odot \mathbf{A},
\end{equation}

\noindent
where redundant gradients are also filtered out and only depthwise weights are activated.

\subsection{Grouping Mechanism}
\label{sec:grouping}

Compared to previous methods for implementing depthwise convolutions,
our method introduces extra computational cost due to the refactorization of weight vectors,
making our method inefficient when the number of input channels is very large.
We propose a grouping mechanism to address this problem,
where
depthwise convolutions are divided into diagonalwise groups and the diagonalwise refactorization is performed for each group.

For a depthwise convolution with $M$ input channels,
the grouping mechanism has the following three steps.
First,
we divide the input channels into $G$ groups, each of which contains $M/G$ channels.
Second,
every group of weight vectors are refactorized into a diagonalwise matrix for that group.
Third,
each group is computed as a standard convolution which supports to leverage the cuDNN library.
By this means
an $M$-channel depthwise convolution is converted into $G$ standard convolutions each having $M/G$ input channels,
instead of one large standard convolution having $M$ input channels.

\subsection{Analysis}

Compared to our diagonalwise refactorization method,
the channel-by-channel method (adopted in Caffe, PyTorch and MXNet)
launches smaller CUDA kernels with fewer threads,
because the workload of a single input channel is small.
The one-channel-at-a-time computation leads to low utilization of GPU resources.
In contrast,
our method rearranges the input channels into diagonalwise groups,
and the computation within a group could be done by a single CUDA kernel or cuDNN function.
This enables the computation of input channels in the same group to be executed in parallel
and thus more GPU threads could be launched.
Consequently,
our method obtains higher computation parallelism and has a higher utilization of GPU resources.

The specialized kernel method (adopted in TensorFlow) launches one specialized CUDA kernel for all the input channels of a depthwise convolution layer,
and achieves considerable improvement in GPU resource utilization compared to the channel-by-channel method.
However, it directly computes the convolutions
and cannot leverage the algorithm-/microarchitecture-level optimizations of the cuDNN library.
In contrast,
our method convert a depthwise convolution into a standard convolution
which supports to leverage the cuDNN library to compute convolutions.
At the algorithm level,
cuDNN provides fast algorithms (such as Fast Fourier Transform and Winograd Transform) for standard convolutions.
At the microarchitecture level,
cuDNN uses kernels specifically optimized for NVIDIA GPUs with full support of shared memory and caches.
Consequently,
our method is more efficient in leveraging the GPU capacity
compared to the specialized kernel method of TensorFlow.

\begin{table*}[ht]
\centering
\scriptsize
\begin{threeparttable}[ht]
\caption{Training Time on MobileNets.}
\label{table:overall-results}
\begin{tabular}{|c|c|ccc|c|ccc|c|}
\hline
\textbf{Frameworks} & \begin{tabular}[c]{@{}c@{}}\textbf{cuDNN}\\\textbf{version}\end{tabular} & \begin{tabular}[c]{@{}c@{}}\textbf{C-by-C}\\\textbf{GEMM}\end{tabular} & \begin{tabular}[c]{@{}c@{}}\textbf{C-by-C}\\\textbf{GEMM Stream}\end{tabular} & \begin{tabular}[c]{@{}c@{}}\textbf{C-by-C}\\\textbf{cuDNN}\end{tabular} & \begin{tabular}[c]{@{}c@{}}\textbf{Specialized}\\\textbf{Kernel}\end{tabular} & \begin{tabular}[c]{@{}c@{}}\textbf{Diagonalwise}\\ \textbf{GEMM}\end{tabular} & \begin{tabular}[c]{@{}c@{}}\textbf{Diagonalwise}\\\textbf{cuDNN (w/o grouping)}\end{tabular} & \begin{tabular}[c]{@{}c@{}}\textbf{Diagonalwise}\\\textbf{cuDNN}\end{tabular} & \begin{tabular}[c]{@{}c@{}}\textbf{Standard}\\\textbf{Convolution}\end{tabular} \\ \hline
\multirow{3}{*}{Darknet} & 5.1 & \multirow{3}{*}{2854.89} & \multirow{3}{*}{2232.09} & 706.28 & \multirow{3}{*}{458.22} & \multirow{3}{*}{627.56} & 343.19 & \textbf{194.21} & 345.09 \\
 & 6.0 &  &  & 864.18 &  &  & 280.70 & \textbf{193.11} & 285.22 \\
 & 7.0 &  &  & 517.06 &  &  & 272.81 & \textbf{185.54} & 275.23 \\ \hline
\multirow{3}{*}{Caffe} & 5.1 & \multirow{3}{*}{2990.05$^*$} & \multirow{3}{*}{-} & 863.63 & \multirow{3}{*}{634.03} & \multirow{3}{*}{777.79} & 500.96 & \textbf{360.39} & 495.99 \\
 & 6.0 &  &  & 1000.19 &  &  & 452.63 & \textbf{357.25} & 450.05 \\
 & 7.0 &  &  & 985.90 &  &  & 452.39 & \textbf{355.69} & 451.37 \\ \hline
PyTorch 0.1.12 & 5.1 & - & - & 816.83$^*$ & - & - & 244.21$^\dag$ & \textbf{152.47}$^\dag$ & 246.72 \\
PyTorch 0.2.0 & 5.1 & - & - & 812.00$^*$ & - & - & 241.61$^\dag$ & \textbf{150.09}$^\dag$ & 244.11 \\ \hline
MXNet 0.10.0 & 5.1 & - & - & 554.16$^*$ & - & - & 211.01$^\dag$ & \textbf{157.09}$^\dag$ & 211.14 \\ \hline
TensorFlow 1.2 & 5.1 & - & - & - & 241.05$^*$ & - & 251.82$^\dag$ & \textbf{177.56}$^\dag$ & 255.85 \\
TensorFlow 1.3 & 6.0 & - & - & - & 243.37$^*$ & - & 253.94$^\dag$ & \textbf{179.59}$^\dag$ & 258.43 \\ \hline
\end{tabular}
\footnotesize
Time is measured in ms/batch with a batch size of 64.
The \textbf{bold} results of \texttt{Diagonalwise cuDNN} surpass all other methods.
$^*$ indicates the original implementation in the framework. All others are our re-implementations.
``-'' indicats the method is not implemented in the framework.
$^\dag$ indicates the method is implemented using Python API and can be further optimized.
\texttt{C-by-C}: channel-by-channel.
\end{threeparttable}
\end{table*}

\section{Experiments}

In this section, we introduce the experimental results
when adopting our diagonalwise refactorization method
in training MobileNets \cite{howard2017mobilenets} on state-of-the-art frameworks.
The results show that our method gains significant acceleration compared to the original implementations in these frameworks.
We further investigate the influence of different grouping strategies and network architectures on our method.
We also evaluate the memory efficiency of our method.
All experiments are conducted on an NVIDIA GTX 1080Ti GPU,
and each result is an average of 1000 batches (with batch size of 64) in the training procedure.

\subsection{Acceleration on MobileNets}

We have implemented our method on five popular deep learning frameworks
including Darknet, Caffe, PyTorch, MXNet and TensorFlow.
The results are shown in Table \ref{table:overall-results},
where ``$*$'' indicates the original implementation in the corresponding framework
(and all other cells in the table are implemented by us).

To separate the effect of our design,
we evaluate three versions of our implementation.
(i) \texttt{Diagonalwise GEMM} represents our implementation that does not utilize the cuDNN library;
(ii) \texttt{Diagonalwise cuDNN (w/o grouping)} represents the implementation that utilizes cuDNN without the grouping mechanism;
and (iii) \texttt{Diagonalwise cuDNN} represents the full implementation that utilizes cuDNN with grouping.
We also implement and evaluate the training performance of four typical methods for depthwise convolutions on Caffe and Darknet,
including three \textit{channel-by-channel} (C-by-C) methods
(\texttt{C-by-C GEMM}, \texttt{C-by-C GEMM Stream} and \texttt{C-by-C cuDNN})
as well as the \textit{specialized kernel} method,
where
\texttt{C-by-C GEMM} represents the channel-by-channel implementation using only GEMM API,
\texttt{C-by-C GEMM Stream} represents all GEMM operations are pipelined with CUDA streams to gain acceleration,
and \texttt{C-by-C cuDNN} represents the channel-by-channel implementation using the cuDNN library.

First,
the specialized kernel methods provide considerably higher efficiency than the channel-by-channel methods,
but have slightly worse or similar performance compared to the standard convolutions.
The \texttt{Diagonalwise GEMM} implementation of our method performs better than all channel-by-channel methods
without using cuDNN or the grouping mechanism,
but worse than the specialized kernel methods.
When cuDNN is utilized (\texttt{Diagonalwise cuDNN w/o grouping}),
our method offers better training performance than most specialized kernel methods except on TensorFlow.
Finally, with a carefully selected grouping strategy
(which will be discussed later),
our method (\texttt{Diagonalwise cuDNN}) surpasses all other methods.

\begin{table*}[t]
\centering
\begin{threeparttable}
\caption{Training Time of Different Depthwise Convolution Layers during Forward and Backward Propagation on Caffe.}
\label{table:layer-by-layer-caffe}
\begin{tabular}{|c|c|ccc|cccc|}
\hline
\multirow{3}{*}{\textbf{Layer Number}} & \multirow{3}{*}{\textbf{Layer Configuration}} & \multicolumn{3}{c|}{\textbf{Forward}} & \multicolumn{4}{c|}{\textbf{Backward}} \\ \cline{3-9}
 & & \begin{tabular}[c]{@{}c@{}}\textit{\textbf{C-by-C}}\\\textit{\textbf{GEMM}}\end{tabular} & \begin{tabular}[c]{@{}c@{}}\textit{\textbf{Specialized}}\\\textit{\textbf{Kernel}}\end{tabular} & \begin{tabular}[c]{@{}c@{}}\textit{\textbf{Diagonalwise}}\\\textit{\textbf{cuDNN}}\end{tabular} & \begin{tabular}[c]{@{}c@{}}\textit{\textbf{C-by-C}}\\\textit{\textbf{GEMM}}\end{tabular} & \begin{tabular}[c]{@{}c@{}}\textit{\textbf{Specialized}}\\\textit{\textbf{Kernel}}\end{tabular} & \begin{tabular}[c]{@{}c@{}}\textit{\textbf{Specialized}}\\\textit{\textbf{Kernel}}$^*$\end{tabular} & \begin{tabular}[c]{@{}c@{}}\textit{\textbf{Diagonalwise}}\\\textit{\textbf{cuDNN}}\end{tabular} \\ \hline
2 & $3 \times 3 / 1$, $112 \times 112 \times 32$ & 19.63 & 10.29 & \textbf{7.45} & 194.55 & 186.79 & 23.22 & \textbf{22.84} \\
4 & $3 \times 3 / 2$, $112 \times 112 \times 64$ & 16.56 & 4.63 & \textbf{3.94} & 109.85 & 46.14 & 13.51 & \textbf{15.93} \\
6 & $3 \times 3 / 1$, $56 \times 56 \times 128$ & 32.03 & 8.51 & \textbf{7.45} & 209.66 & 81.44 & 17.47 & \textbf{16.50} \\
8 & $3 \times 3 / 2$, $56 \times 56 \times 128$ & 22.72 & 2.06 & \textbf{2.05} & 78.16 & 12.73 & 6.59 & \textbf{7.02} \\
10 & $3 \times 3 / 1$, $28 \times 28 \times 256$ & 45.54 & 3.90 & \textbf{3.86} & 150.51 & 20.74 & 8.42 & \textbf{7.79} \\
12 & $3 \times 3 / 2$, $28 \times 28 \times 256$ & 37.17 & \textbf{1.06} & 1.19 & 75.52 & 5.22 & 3.40 & \textbf{3.86} \\
14, 16, 18, 20, 22 & $3 \times 3 / 1$, $14 \times 14 \times 512$ & 79.31 & \textbf{2.00} & 2.06 & 159.60 & 7.21 & 4.21 & \textbf{4.60} \\
24 & $3 \times 3 / 2$, $14 \times 14 \times 512$ & 69.42 & \textbf{0.59} & 1.10 & 133.97 & \textbf{2.32} & 1.84 & 4.41 \\
26 & $3 \times 3 / 1$, $7 \times 7 \times 1024$ & 139.03 & \textbf{1.06} & 1.50 & 268.62 & \textbf{3.24} & 2.28 & 3.49 \\ \hline
\multicolumn{2}{|c|}{Total} & 461.42 & 34.10 & 30.61 & 1380.43 & 365.84 & 80.95 & 86.44 \\ \hline
\end{tabular}
Time is measured in ms/batch with a batch size of 64.
\texttt{C-by-C GEMM} is the original implementation in Caffe.
\texttt{Diagonalwise cuDNN} utilizes cuDNN and is grouped with group size of 32.
\texttt{Layer Configuration} shows kernel size, stride and size of the input feature map.
$^*$ indicates the gradients of weights are not computed.
\end{threeparttable}
\end{table*}

Second,
our method achieves $15.4\times$ speedup on Darknet, $8.4\times$ on Caffe, $5.4\times$ on PyTorch and $3.5\times$ on MXNet compared to their \textit{channel-by-channel} methods,
and achieves $1.7\times$ speedup on Darknet and $1.4\times$ speedup on Caffe and TensorFlow
over their \textit{specialized kernel} methods.
The overall speedup could be decomposed to different parts of our method.
On Caffe,
for example,
\texttt{Diagonalwise GEMM} offers $3.84 \times$ speedup over the original \texttt{C-by-C GEMM},
while \texttt{Diagonalwise cuDNN w/o grouping} and \texttt{Diagonalwise cuDNN} incrementally offer $1.72 \times$ and $1.27 \times$ speedups,
resulting in an overall speedup of $8.4$ $(\approx 3.84 \times 1.72 \times 1.27)$ times.
The ablation analysis shows that in our method the refactorization contributes the most in the speedup.
Note that the implementations of our method on PyTorch, MXNet and TensorFlow use Python API provided by the frameworks,
meaning that they do not benefit from operator fusion or simplification
and thus could be further accelerated when being implemented in C++.

\subsection{Layer-by-layer Experiments}

To make a detailed comparison of these methods, we further conduct layer-by-layer experiments evaluating the time of the three different training methods on Caffe.

From Table \ref{table:layer-by-layer-caffe},
the \textit{channel-by-channel} method (\texttt{C-by-C GEMM}) has very poor performance in the last few layers.
These layers all have wide-and-small feature maps
and have at least 256 input channels.
None of them has feature maps larger than $28 \times 28$.
For \texttt{C-by-C GEMM},
only a small number of threads are launched for small feature maps
and the utilization of GPU resources is low.
For instance,
the last depthwise convolution layer (No. 26) of \texttt{C-by-C GEMM}
has only $1/8$ threads compared to the first row of layer (No. 2).
This amplifies the performance gap between the channel-by-channel method and other methods.

During forward propagation,
\texttt{Diagonalwise cuDNN} outperforms all other methods in the total training time.
\texttt{Diagonalwise cuDNN} is better than \texttt{Specialized Kernel} in the first layers,
but \texttt{Specialized Kernel}
surpasses \texttt{Diagonalwise cuDNN} in the last few layers.
This is because
small feature maps usually lead to more efficient usage of shared memory and higher cache-hit rate,
which improve the performance of the specialized kernel method.
For \texttt{Diagonalwise cuDNN},
it is relatively difficult to find a balance between computational redundancy and GPU utilization
when the number of channels is large,
resulting in slightly lower performance compared to the specialized kernel method
in the last layers.

\begin{table}[t]
\centering
\begin{threeparttable}
\caption{Training Time of Different Layer Types on Caffe.}
\label{table:layer-type-2}
\begin{tabular}{|c|c|c|c|}
\hline
\textbf{Type} & \begin{tabular}[c]{@{}c@{}}\textbf{C-by-C}\\\textbf{GEMM}\end{tabular} & \begin{tabular}[c]{@{}c@{}}\textbf{Specialized}\\\textbf{Kernel}\end{tabular} & \begin{tabular}[c]{@{}c@{}}\textbf{Diagonalwise}\\\textbf{cuDNN}\end{tabular} \\ \hline
Conv $1 \times 1$ & 16.39\% & 24.02\% & 46.43\% \\ \hline
Conv DW $3 \times 3$ & 82.86\% & 71.79\% & 45.41\% \\ \hline
Conv $3 \times 3$ & 0.72\% & 4.04\% & 7.87\% \\ \hline
Fully Connected & 0.03\% & 0.15\% & 0.29\% \\ \hline
\end{tabular}
\texttt{Diagonalwise cuDNN} utilizes cuDNN and is grouped with group size of 32.
\texttt{C-by-C}: channel-by-channel.
\end{threeparttable}
\end{table}

During backward propagation,
\texttt{Diagonalwise cuDNN} significantly outperforms other methods.
To understand the poor performance of \texttt{Specialized Kernel},
we also evaluate it without computing the gradients of the weights (\texttt{Specialized Kernel$^*$}).
The result shows that \texttt{Specialized Kernel} spends most of its training time in computing the gradients.
This is because there are memory write hazards when computing the gradients,
and the atomic operations slow down the whole training procedure of the specialized kernel method.
This problem is more serious in the first few layers,
where the feature maps are larger and memory write conflicts are more frequent.
Optimized kernels and algorithms in the cuDNN library give significant acceleration to our diagonalwise refactorization method.

\begin{table*}[t]
\centering
\begin{threeparttable}
\caption{Training Time of MobileNet Variants with Different Hyperparameters.}
\label{table:variants-result}
\begin{tabular}{|c|c|cc|c|cc|}
\hline
\textbf{Architecture} & \textbf{Framework} & \textbf{C-by-C GEMM} & \textbf{C-by-C cuDNN} & \textbf{Specialized Kernel} & \textbf{Diagonalwise cuDNN} & \textbf{Grouping Strategy} \\ \hline
\multirow{5}{*}{\begin{tabular}[c]{@{}c@{}}Shallow\\ MobileNet\end{tabular}} & Darknet & 1644.92 & 486.63 & 412.20 & \textbf{151.12} & \multirow{4}{*}{size=32} \\
 & Caffe & 1776.58 & 600.66 & 545.14 & \textbf{274.55} &  \\
 & PyTorch 0.2.0 & - & 492.84$^*$ & - & \textbf{116.34}$^\dag$ &  \\
 & MXNet 0.10.0 & - & 361.45$^*$ & - & \textbf{116.51}$^\dag$ &  \\
 & TensorFlow 1.2 & - & - & 206.90$^*$ & \textbf{137.46}$^\dag$ & size=64 \\ \hline
\multirow{5}{*}{\begin{tabular}[c]{@{}c@{}}Thinner\\MobileNet\\(Width\\ Multiplier\\ 0.75)\end{tabular}} & Darknet & 2121.06 & 524.10 & 347.81 & \textbf{172.17} & \multirow{4}{*}{size=24} \\
 & Caffe & 2269.27 & 670.91 & 490.56 & \textbf{295.19} &  \\
 & PyTorch 0.2.0 & - & 605.23$^*$ & - & \textbf{110.78}$^\dag$ &  \\
 & MXNet 0.10.0 & - & 415.97$^*$ & - & \textbf{118.51}$^\dag$ &  \\
 & TensorFlow 1.2 & - & - & 209.35$^*$ & \textbf{141.07}$^\dag$ & size=48 \\ \hline
\multirow{5}{*}{\begin{tabular}[c]{@{}c@{}}Thinner\\MobileNet\\(Width\\ Multiplier\\ 0.5)\end{tabular}} & Darknet & 1358.60 & 341.69 & 230.39 & \textbf{102.14} & \multirow{4}{*}{size=32} \\
 & Caffe & 1473.85 & 465.80 & 343.40 & \textbf{209.50} &  \\
 & PyTorch 0.2.0 & - & 396.86$^*$ & - & \textbf{74.06}$^\dag$ &  \\
 & MXNet 0.10.0 & - & 280.69$^*$ & - & \textbf{84.42}$^\dag$ &  \\
 & TensorFlow 1.2 & - & - & 155.56$^*$ & \textbf{85.50}$^\dag$ & size=64 \\ \hline
\multirow{5}{*}{\begin{tabular}[c]{@{}c@{}}Low-resolution\\MobileNet\\(Resolution\\ Multiplier\\ 128)\end{tabular}} & Darknet & 2219.51 & 443.16 & 97.25 & \textbf{69.25} & \multirow{4}{*}{size=32} \\
 & Caffe & 2201.68 & 544.13 & 170.35 & \textbf{150.88} &  \\
 & PyTorch 0.2.0 & - & 410.85$^*$ & - & \textbf{54.83}$^\dag$ &  \\
 & MXNet 0.10.0 & - & 280.38$^*$ & - & \textbf{56.46}$^\dag$ &  \\
 & Tensorflow 1.2 & - & - & 78.98$^*$ & \textbf{65.83}$^\dag$ & size=64 \\ \hline
\end{tabular}
Time is measured in ms/batch with a batch size of 64.
\texttt{Diagonalwise cuDNN} utilizes cuDNN and is grouped with group size of 32.
\texttt{C-by-C}: channel-by-channel.
The \textbf{bold} results of \texttt{Diagonalwise cuDNN} surpass all other methods.
$^*$ indicates the original implementation in the framework.
$^\dag$ indicates the method is implemented using Python API.
cuDNN 5.1 is used in all frameworks.
\end{threeparttable}
\end{table*}

Table \ref{table:layer-type-2} shows the training time of each layer type using three methods to train a standard MobileNet.
Compared to \texttt{C-by-C GEMM} and \texttt{Specialized Kernel},
the ratio of time spent on training depthwise convolution layers
dramatically decreases (to $45.41\%$)
using the diagonalwise refactorization method (\texttt{Diagonalwise cuDNN}).

\begin{figure}[t]
\centering
\subfigure[]{
\includegraphics[width=0.46\textwidth]{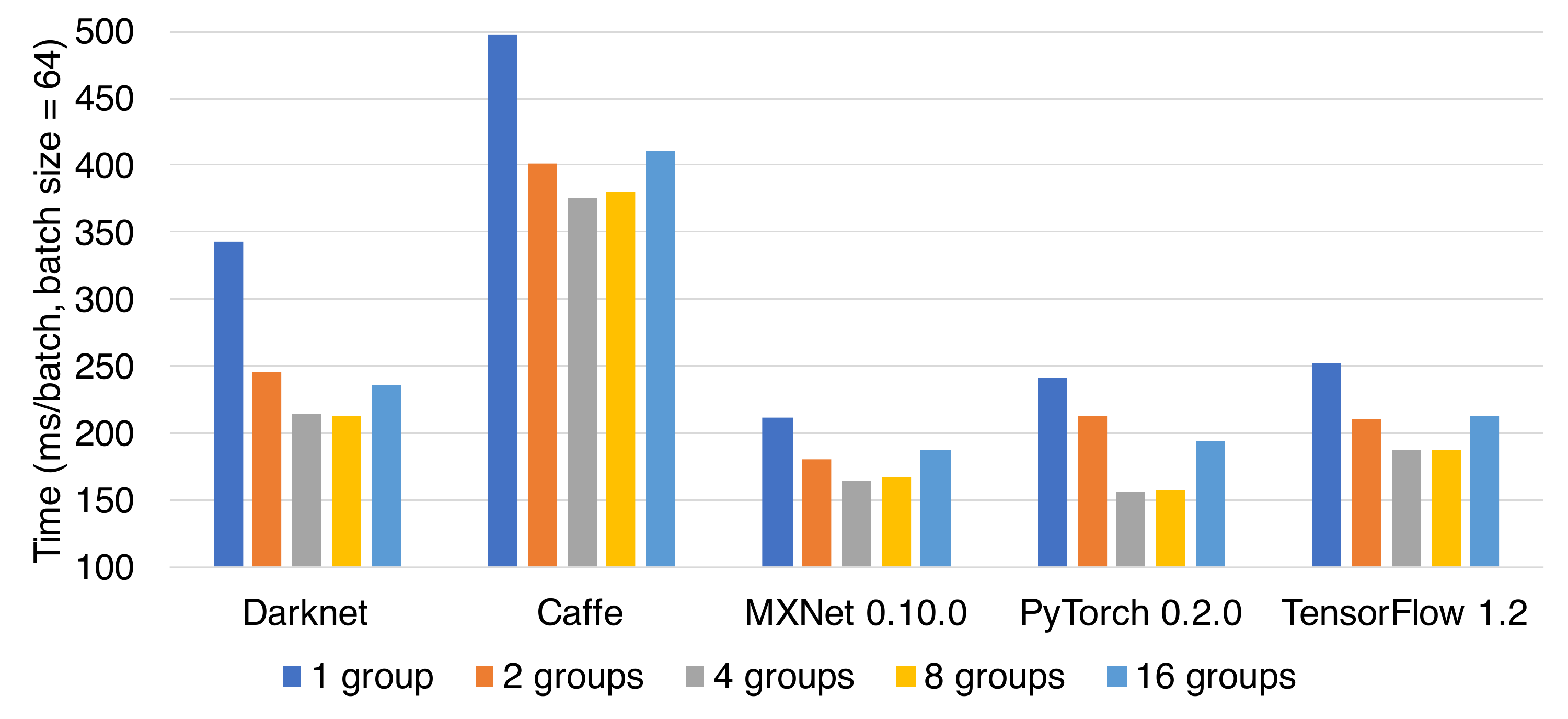}
\label{figure:group-by-number}
}
\subfigure[]{
\includegraphics[width=0.46\textwidth]{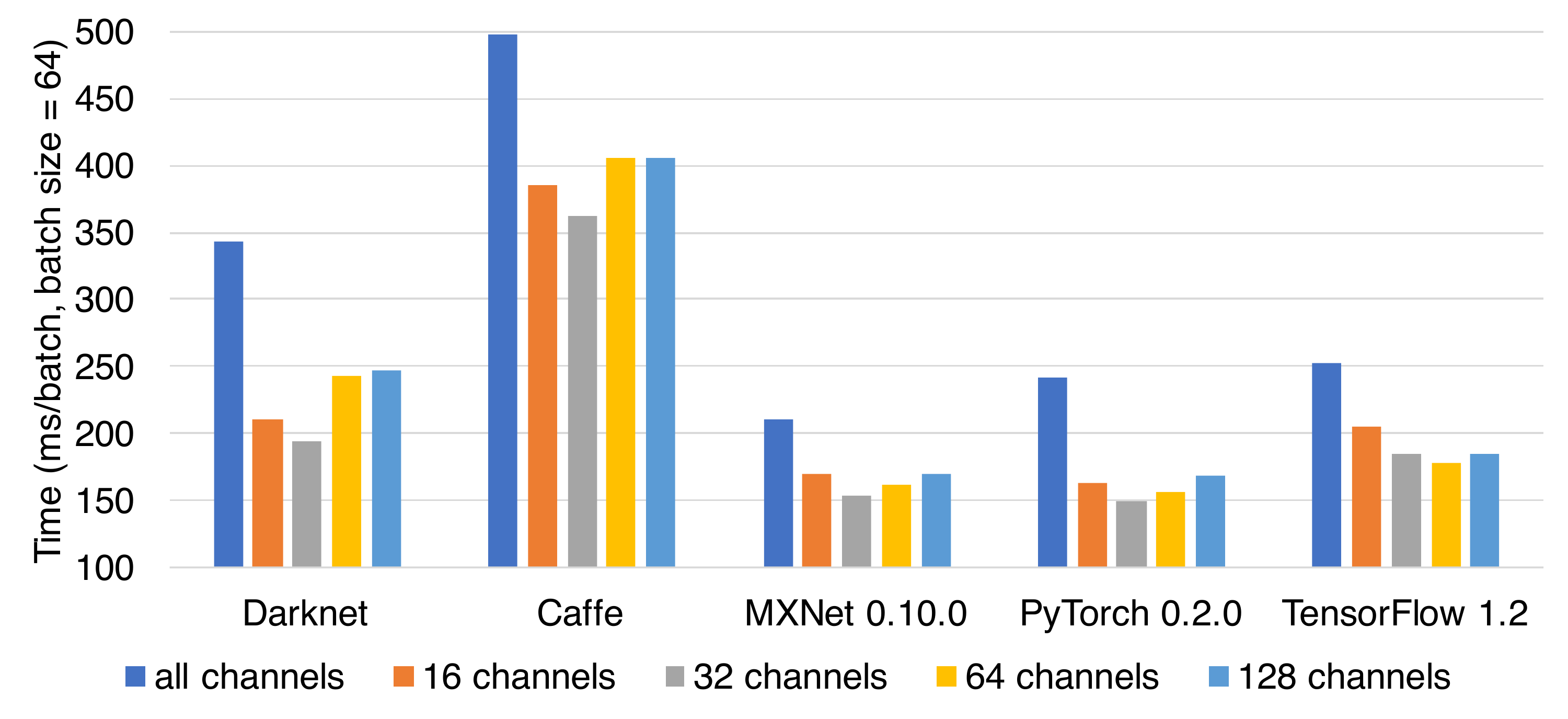}
\label{figure:group-by-size}
}
\caption{Training time with (a) grouping by the number of groups and (b) grouping by the group size. cuDNN 5.1 is used in all frameworks. Implementations on PyTorch, MXNet and TensorFlow use Python API.}
\label{figure:group-strategy}
\end{figure}

\subsection{Grouping Strategies}

We propose two strategies for grouping in our method
and investigate their influence on the training performance.

\noindent \textbf{Grouping by the number}.
The first strategy is to group by the number of groups.
Assuming that the number of groups is $G$, every depthwise convolution layer is divided into $G$ diagonalwise groups and each group contains $M/G$ channels.
We compare the efficiency of our method with $G$ from 1 to 16 and the results are demonstrated in Figure \ref{figure:group-by-number}.

\noindent \textbf{Grouping by the size}.
The second strategy is to group by the group size.
Assuming that the group size is $S$ channels, there will be $M/S$ groups in each depthwise convolution layer after grouping.
We compare the efficiency of our method with $S$ from 16 to 128.
Figure \ref{figure:group-by-size} shows the results under different configurations.

From Figure \ref{figure:group-strategy},
the utilization of a group strategy considerably improves the performance of our method.
Compared to the methods without grouping,
grouping by the number offers $1.6 \times$ speedup
while grouping by the size achieves $1.76 \times$ speedup.
Grouping by the size outperforms grouping by the number,
because the first few thin layers does not suffer from the redundant computation.

\subsection{Different MobileNets Hyperparameters}

To evaluate the generalization ability of our method,
we conduct extensive experiments on MobileNet variants with different hypermeters.
The variants include shallow MobileNet, thinner MobileNet with width multiplier of 0.75 and 0.5, and low-resolution MobileNet with resolution multiplier of 128,
respectively representing networks with fewer layers, fewer channels and smaller feature maps.
Table \ref{table:variants-result} shows the results for these variants
where
\texttt{Diagonalwise cuDNN} significantly outperforms all other methods,
demonstrating our method is adaptive to different networks built with depthwise convolutions.

\begin{table}[!ht]
\centering
\begin{threeparttable}
\caption{Memory Consumption on Darknet, Caffe and PyTorch.}
\label{table:memory-consumption}
\begin{tabular}{|c|cc|c|c|}
\hline
\textbf{Framework} & \begin{tabular}[c]{@{}c@{}}\textbf{C-by-C}\\\textbf{GEMM}\end{tabular} & \begin{tabular}[c]{@{}c@{}}\textbf{C-by-C}\\\textbf{cuDNN}\end{tabular} & \begin{tabular}[c]{@{}c@{}}\textbf{Specialized}\\\textbf{Kernel}\end{tabular} & \begin{tabular}[c]{@{}c@{}}\textbf{Diagonalwise}\\\textbf{cuDNN}\end{tabular} \\ \hline
Darknet & 5369MB & 5355MB & 5355MB & 5355MB \\
Caffe & 8015MB & 7973MB & 7947MB & 7981MB \\
PyTorch 0.2.0 & - & 3795MB & - & 3807MB \\ 
\hline
\end{tabular}
\texttt{C-by-C GEMM} is the original implementation in Caffe.
\texttt{Diagonalwise cuDNN} utilizes cuDNN and is grouped with group size of 32.
cuDNN 5.1 is used in all frameworks.
\end{threeparttable}
\end{table}

Compared to the results on standard MobileNets (Table \ref{table:overall-results}),
our method offers lower speedup over \textit{channel-by-channel} methods
but higher speedup over \textit{specialized kernel} methods on shallow MobileNet and thinner MobileNet:
our method offers $6.47 \times$ speedup on Caffe and $1.8 \times$ on TensorFlow over their original implementations.
But on low-resolution MobileNets, the results are opposite:
our method achieves $14.6 \times$ speedup on Caffe and only $1.2 \times$ on TensorFlow.
These differences are attributed to the change in the number of wide-and-small depthwise convolution layers.
Shallow MobileNet removes 5 wide-and-small depthwise convolution layers,
while thinner MobileNet reduce the number of channels in every layer.
This reduces the number of wide-and-small layers, which improves the average utilization of GPU resources in the channel-by-channel methods and provides better performance.
But for the specialized kernel methods, the average shared memory and cache hit-rate is lower.
Low-resolution MobileNet shrink every feature map in the networks.
Small features maps further reduce the utilization of GPU resources in the channel-by-channel methods.
The specialized kernel methods benefit more from the shrinking of feature maps because of higher shared memory and caches hit rate.

\subsection{Memory Efficiency}
\label{sec:memory}

One potential side effect of our method is that the increase of the parameters may lead to large GPU memory usage.
We present extensive experiments with GPU memory usage on Darknet, Caffe and PyTorch to evaluate the memory efficiency of our method.
We compare our method with two channel-by-channel methods (\texttt{C-by-C GEMM} and \texttt{C-by-C cuDNN}) and the specialized kernel method.
From Table \ref{table:memory-consumption},
the increase in memory usage of our method is negligible compared to \texttt{C-by-C cuDNN} and the specialized kernel method,
and our method consumes less GPU memory than \texttt{C-by-C GEMM},
proving our method is memory-efficient.

\section{Discussion}

\subsection{Extra Computational Cost}

As demonstrated in Section \ref{sec:grouping},
by converting a depthwise convolution into a standard convolution using diagonalwise refactorization,
extra computational cost is introduced.
But this extra computational cost can be ignored compared with the speedup achieved by our method for three reasons.

Firstly, our method is focused on the \textit{training} performance of ConvNets
built with depthwise convolutions.
During training procedure, execution speed and memory consumption are the two key points
which affect the training time and the batch size,
while the number of floating point operations is less concerned.
From these two aspects, 
our method achieves significant speedup on five popular deep learning frameworks
with little impact on memory utilization (as demonstrated in Section \ref{sec:memory}).
These advantages make our method more efficient for training depthwise convolutions.

Secondly, in practice, the reduction in computational cost by depthwise convolutions does not provide high training speed
on hardware with high parallelism like GPUs.
This is because the performance is significantly affected by the hit rates of shared memory and cache,
but not rigorously related to the number of floating point operations.
From Table \ref{table:layer-type-1},
channel-by-channel method and specialized kernel method spend most of the training time (83\% for channel-by-channel method and 72\% for specialized kernel method) on depthwise convolutions,
which have only 3\% of the FLOPs,
due to the inefficient shared memory utilization (in channel-by-channel method)
and the lack of algorithm-level and microarchitecture-level optimization (in specialized kernel method).
Instead, our method increases the inter-channel parallelism and leverages the highly efficient cuDNN optimization.
By this means, the extra cost is hidden by the parallel computation and has little impact on training speed,
which provides significant speedup compared with the other two methods.

Thirdly, our method is much more implementation-friendly than the specialized kernel method of TensorFlow,
and provides better compatibility by using cuDNN.
Instead of manually designing kernels to handle various cases,
our method only needs to add a simple mask filtering on regular convolutions and provide high efficiency.

\subsection{Extensibility}

In this subsection
we briefly discuss the extensibility of the proposed method.
The utilization of mask matrices makes diagonalwise refactorization
easy to be adopted in networks with sparse connections.

One example is group convolution.
Group convolutions are adopted in the state-of-the-art networks such as \cite{xie2016aggregated} and \cite{zhang2017shufflenet}.
Depthwise convolution could be viewed as a special case of group convolution
where the number of groups equals the number of channels.
In a more general case,
several groups in a group convolution can be rearranged into a larger diagonalwise group
using our method
and the computation can be accelerated by increasing the parallelism between groups.
Moreover, our method can be utilized to more flexible networks architectures which adopt convolutional kernels of different sizes to each group.

Another example is network pruning,
a technique that cuts off redundant weights/connections and accelerates the inference procedure.
The flexibility of network pruning strategies introduces extra cost into the training procedure,
because the connections between neurons in a network become sparser and harder to control.
Our method can accelerate the training procedure and eliminate the extra cost 
by setting pruned positions in mask matrices to 0 and other positions to 1,
and offers significant speedup when pruning networks built with depthwise convolutions and group convolutions.

\section{Conclusion}

In this paper,
we analyze the problems of typical methods for implementing depthwise convolutions in popular deep learning frameworks,
and propose a novel method (called diagonalwise refactorization)
to accelerate the training of depthwise convolution layers.
By rearranging the depthwise filters in a large diagonal weight matrix,
our method increases the computation parallelism in depthwise convolutions.
Experiments on five popular frameworks show that
the diagonalwise refactorization method offers significant acceleration.

\section*{Acknowledgment}

This work is supported by the Major State Basic Research Development Program of China (973) under Grant no. 2014CB340303.

\bibliographystyle{IEEEtran}
\bibliography{main}

\end{document}